\documentclass[pdflatex,sn-mathphys-num]{sn-jnl}

\usepackage{graphicx}%
\usepackage{amsmath,amssymb,amsfonts}%
\usepackage{amsthm}%
\usepackage{mathrsfs}%
\usepackage[title]{appendix}%
\usepackage{xcolor}%
\usepackage{textcomp}%
\usepackage{manyfoot}%
\usepackage{booktabs}%
\usepackage{algorithm}%
\usepackage{algorithmicx}%
\usepackage{algpseudocode}%
\usepackage{listings}%
\usepackage{caption}%
\usepackage{multirow}

\newcommand{\quotes}[1]{``#1''}

\theoremstyle{thmstyleone}%
\theoremstyle{thmstyletwo}%

\theoremstyle{thmstylethree}%

\begin{document}

\title[Multi-task Learning for Identification of Porcelain]{Multi-task Learning for Identification of Porcelain in Song and Yuan Dynasties}


\author[1]{\fnm{Ziyao} \sur{Ling}}\email{ziyao.ling2@unibo.it}

\author*[1]{\fnm{Giovanni} \sur{Delnevo}}\email{giovanni.delnevo@unibo.it}

\author[1]{\fnm{Paola} \sur{Salomoni}}\email{paola.salomoni@unibo.it}

\author[1]{\fnm{Silvia} \sur{Mirri}}\email{silvia.mirri@unibo.it}

\affil*[1]{\orgdiv{Department of Computer Science and Engineering}, \orgname{University of Bologna}, \orgaddress{\street{Mura Anteo Zamboni 7}, \city{Bologna}, \postcode{40127}, \state{Emilia-Romagna}, \country{Italy}}}


\abstract{Chinese porcelain holds immense historical and cultural value, making its accurate classification essential for archaeological research and cultural heritage preservation. Traditional classification methods rely heavily on expert analysis, which is time-consuming, subjective, and difficult to scale. This paper explores the application of DL and transfer learning techniques to automate the classification of porcelain artifacts across four key attributes: dynasty, glaze, ware, and type. We evaluate four Convolutional Neural Networks (CNNs) — ResNet50, MobileNetV2, VGG16, and InceptionV3 — comparing their performance with and without pre-trained weights. Our results demonstrate that transfer learning significantly enhances classification accuracy, particularly for complex tasks like type classification, where models trained from scratch exhibit lower performance. MobileNetV2 and ResNet50 consistently achieve high accuracy and robustness across all tasks, while VGG16 struggles with more diverse classifications. We further discuss the impact of dataset limitations and propose future directions, including domain-specific pre-training, integration of attention mechanisms, explainable AI methods, and generalization to other cultural artifacts..}




\maketitle

\section{Introduction}\label{sec:intro}

Chinese porcelain holds an esteemed place in both cultural heritage and art history, renowned for its craftsmanship, aesthetic value, and historical importance \cite{xu2022cultural}. Dating back to the Tang and Song dynasties, porcelain artifacts reflect the evolution of techniques, glazes, and decorative styles over centuries. Accurately classifying these artifacts into categories such as dynasty, glaze, ware, and type is crucial for historians, archaeologists, and collectors, as it aids in provenance research, forgery detection, and understanding historical trade routes \cite{finlay2010pilgrim}. Traditional classification methods rely heavily on expert knowledge and manual inspection, which are time-consuming and prone to subjectivity \cite{di2022review}.

The application of Artificial Intelligence (AI) within the cultural heritage domain has emerged as a transformative tool, offering opportunities to augment traditional methods. AI-driven analysis can enhance artifact identification, restoration, and preservation efforts by processing vast collections rapidly and uncovering patterns that may elude human experts \cite{prados2023new, liu2023unsupervised, girbacia2024analysis}. For instance, Deep Learning (DL) techniques have been employed to restore damaged frescoes \cite{merizzi2024deep} and authenticate paintings \cite{elgammal2018shape}. These successes demonstrate AI's potential to revolutionize how we engage with and protect cultural heritage, making it more accessible to researchers and the public alike.

Machine Learning (ML) algorithms have emerged as powerful tools not only for automating and enhancing artifact classification \cite{ling2024findings} but also from a digital sustainability perspective. By digitizing and automating the classification process, this approach reduces the need for repeated physical handling of fragile artifacts, minimizing the risk of damage and ensuring long-term preservation of cultural heritage \cite{mazzetto2024integrating}. Furthermore, creating digital repositories of classified artifacts facilitates global accessibility, enabling researchers and institutions to collaborate and study porcelain collections without the environmental and financial costs associated with physical transportation. Such advancements align with sustainable digitization initiatives, fostering a balance between technological innovation and the responsible stewardship of cultural resources for future generations \cite{nag2024sustainable}. Convolutional Neural Networks (CNNs), in particular, have demonstrated remarkable performance in visual recognition tasks \cite{krizhevsky2017imagenet}, making them well-suited for analyzing porcelain imagery. By leveraging pre-trained models such as InceptionV3 \cite{szegedy2016rethinking}, MobileNetV2 \cite{sandler2018mobilenetv2}, ResNet50 \cite{he2016deep}, and VGG16 \cite{simonyan2015very}, researchers can benefit from transfer learning — utilizing knowledge from large-scale datasets like ImageNet \cite{deng2009imagenet} to improve performance on smaller, domain-specific datasets. This approach reduces training time and mitigates the need for extensive labeled data, which is often scarce in cultural heritage studies.

This paper explores the performance of these CNN architectures in classifying porcelain artifacts across multiple tasks: dynasty, glaze, ware, and type. It presents a comparative analysis of models, evaluating their accuracy, balanced accuracy, precision, recall, and F1 score. Furthermore, the impact of transfer learning is investigated by comparing models trained with and without pre-trained weights, offering insights into how domain adaptation affects performance.

The remainder of this paper is structured as follows. Section \ref{sec:related} outlines the related works while Section \ref{sec:met} details the dataset and preprocessing step, the models and experimental settings. Section \ref{sec:res} presents the results and discusses model performance, including a dedicated analysis of transfer learning’s contribution. Section \ref{sec:conc} concludes with key findings and proposes directions for future research, such as integrating attention mechanisms, domain-specific pre-training datasets, and expanding the approach to other cultural artifacts.

\section{Related Work}\label{sec:related}

The following Subsections provide a comprehensive overview of existing methodologies for porcelain classification, covering traditional ML and DL approaches, the role of multi-task and transfer learning in improving performance, and the challenges associated with dataset limitations in archaeological porcelain identification.

\subsection{Automatic Porcelain Identification}

Recent advancements in AI have opened new avenues for the intelligent identification of archaeological porcelains. Given the cultural significance of these artifacts, accurately identifying and classifying them is crucial for historical analysis and preservation. AI-driven methods for porcelain identification can be broadly categorized into ML-based and DL-based approaches. Due to the delicate and valuable nature of archaeological porcelains, these methods predominantly rely on digital images rather than physical handling of the artifacts.

Traditional ML techniques have been widely employed for porcelain classification, primarily through image classification algorithms such as Support Vector Machines (SVM) \cite{zhang2020characteristics} and Mean Shift Clustering \cite{zhou2017porcelain}. These methods extract features from porcelain images to train models capable of recognizing patterns. Studies utilizing ML methods have achieved classification accuracies exceeding 75\%, though these models were often trained on relatively small datasets. While ML approaches are computationally efficient, they rely heavily on hand-crafted feature extraction, which limits their ability to handle the intricate visual details of porcelain artifacts.

More recent studies have demonstrated the effectiveness of DL, particularly CNNs, in the feature extraction and classification of porcelain images. Prominent architectures such as ResNet \cite{yue2023comparative} and VGG16 \cite{ma2021identification} have achieved classification accuracies surpassing 85\%. Unlike ML, DL models automatically learn hierarchical features directly from raw images, capturing complex visual characteristics like texture, glaze patterns, and shapes. However, DL methods demand substantial computational resources and large datasets, which can be a challenge in the context of cultural heritage research.

Complementary to ML and DL approaches, classical image processing techniques — including Local Binary Patterns (LBP) \cite{sun2023identification}, Histogram of Oriented Gradient (HOG) \cite{zhang2020characteristics}, and Gray Level Co-occurrence Matrix (GLCM) \cite{rasheed2015using} — are commonly employed for image enhancement, segmentation, and preprocessing. These methods help refine the visual data, improving feature extraction and ultimately boosting classification accuracy. When combined with ML or DL, image processing enhances model robustness, particularly for artifacts with subtle or degraded features.

\subsection{Multi-Task and Transfer Learning}

Multi-task learning leverages shared representations across multiple related tasks, leading to improved data efficiency, faster learning, and better generalization \cite{crawshaw2020multi}. This technique is particularly beneficial for archaeological porcelain classification, where multiple attributes — such as dynasty, ware, glaze, and type — can be learned concurrently from the same dataset. By sharing feature representations during training, the model optimizes performance across all classification tasks simultaneously.

Transfer learning has emerged as a powerful approach to address the challenges posed by limited datasets. Studies by Liu et al. \cite{liu2022automatic} and Yang et al. \cite{yang2022ceramic} demonstrated the effectiveness of transfer learning using CNN models pre-trained on large datasets like ImageNet. This strategy exploits the general feature representations learned from extensive image datasets, significantly reducing the need for large labeled datasets in specialized domains like porcelain classification. Tajbakhsh et al. \cite{tajbakhsh2016convolutional} further highlighted that the early layers of CNNs learn generic patterns — such as edges and textures — which are transferable across diverse image domains.

Two primary transfer learning strategies are prevalent: feature extraction, where pre-trained models serve as fixed feature extractors, and fine-tuning, where pre-trained weights are partially or fully retrained on the target dataset. This study explores both approaches using four state-of-the-art architectures: ResNet50, MobileNetV2, VGG16, and InceptionV3.

\subsection{Dataset for Porcelain Identification}

Despite the promising results achieved by prior studies, dataset limitations remain a significant bottleneck in archaeological porcelain classification. Weng et al. \cite{weng2017machine} demonstrated high accuracy using a dataset of only 50 images — 10 images per age across five dynasties — highlighting the small scale of many existing datasets. Such datasets often lead to overfitting, reduced generalization, and poor performance on unseen data.

Moreover, many studies restrict feature selection to narrow characteristics. For example, Weng et al. \cite{weng2017machine} focused on porcelain of the same shape from different dynasties, while Zhang et al. \cite{zhang2020characteristics} examined artifacts with similar patterns. This reductionist approach overlooks the rich diversity of Chinese archaeological porcelain, which varies significantly in glaze, texture, composition, and form due to differences in raw materials, production methods, and firing techniques \cite{Shi2005Encyclopedia}. To achieve more comprehensive and reliable classification, future datasets must incorporate diverse features representative of the full spectrum of porcelain artifacts, fostering robust and generalizable model performance.

\section{Materials and Methods}
\label{sec:met}

This Section details the methodology employed in our research. We outline the specific research questions addressed, provide a comprehensive description of the dataset utilized, and elaborate on the multi-task learning framework developed. Furthermore, we define the loss function and evaluation metrics used to assess performance, and conclude with a thorough explanation of the experimental settings.

\subsection{Research Questions}

This work tackle the problem of porcelain artifacts classification, addressing the following research questions:
\begin{itemize}
    \item \textbf{RQ1} \quotes{What level of accuracy can be achieved when classifying the dynasty, glaze, ware, and type of porcelain artifacts using DL techniques?}
    \item \textbf{RQ2} \quotes{What is the impact of transfer learning, using pre-trained CNNs, on the accuracy of classifying dynasty, glaze, ware, and type of porcelain artifacts?}

\end{itemize}

\subsection{Dataset Description}

Sample images of porcelains are collected from the open data of the National Palace Museum in Taipei \cite{national_palace_museum_opendata}. A total of 5,993 porcelain images from the Song (ca. 1064 to 745 aBP) and Yuan (ca. 753 to 565 aBP) dynasties are selected, including different glaze colours, wares and porcelain types. The LabelU tool \cite{he2024opendatalab} was used to manually accurately annotate all images according to the naming standards of the National Palace Museum in Taipei, each image has four labels, that cover the four main porcelain features. The values for each label, together with the corresponding cardinality are reported in Table \ref{tab:labels_distribution}. From the table, it is clear the unbalanced nature of our dataset for each label.

\begin{table*}[h]
\centering
\caption{Porcelain Dataset Distribution among Labels}
\label{tab:labels_distribution}
\begin{tabular}{|l|l|l|l|l|l|l|l|}
\hline
\multicolumn{1}{|c|}{\textbf{Dinasty}} & \multicolumn{1}{c|}{5993} & \multicolumn{1}{c|}{\textbf{Ware}} & \multicolumn{1}{c|}{5993} & \multicolumn{1}{c|}{\textbf{Glaze}} & \multicolumn{1}{c|}{5993} & \multicolumn{1}{c|}{\textbf{Type}} & \multicolumn{1}{c|}{5993} \\ \hline
Song                                   & 5288                      & Ding                               & 2296                      & White                               & 2668                      & Washer                             & 747                       \\ \hline
Yuan                                   & 705                       & Jizhou                             & 167                       & Black                               & 113                       & Dish                               & 1610                      \\ \hline
                                       &                           & Ge                                 & 416                       & Celadon                             & 2379                      & Bowl                               & 2002                      \\ \hline
                                       &                           & Guan                               & 1401                      & Green                               & 577                       & Plate                              & 127                       \\ \hline
                                       &                           & Jun                                & 263                       & Moonwhite                           & 54                        & Teabowlstand                       & 64                        \\ \hline
                                       &                           & Longquan                           & 385                       & Yellowishgreen                      & 4                         & Pillow                             & 112                       \\ \hline
                                       &                           & Ru                                 & 677                       & Bluishwhite                         & 64                        & Basin                              & 244                       \\ \hline
                                       &                           & Xianghu                            & 76                        & Blue                                & 134                       & Vase                               & 565                       \\ \hline
                                       &                           & Linchuan                           & 98                        &                                     &                           & Jar                                & 74                        \\ \hline
                                       &                           & Peng                               & 214                       &                                     &                           & Incenseburner                      & 157                       \\ \hline
                                       &                           &                                    &                           &                                     &                           & Vessel                             & 147                       \\ \hline
                                       &                           &                                    &                           &                                     &                           & Cup                                & 144                       \\ \hline
\end{tabular}
\end{table*}

The Dynasty label distinguishes between Song and Yuan porcelains. The Song dynasty marked a golden age for Chinese porcelain, characterized by distinct ware systems such as Ding (white porcelain), Jun (white and underglaze-colored porcelain), Yaozhou (red and blue glazes), Yue and Longquan (celadon), and Black porcelain. Additionally, official wares such as Ru, Jun, and Bianjing were produced under imperial patronage \cite{Shi2005Encyclopedia}. The Yuan dynasty continued these traditions while innovating with new techniques. While Longquan celadon production remained prevalent, the late Yuan period saw the emergence of blue-and-white, copper red, and other novel glazes in Jingdezhen, shaping the foundation for later porcelain advancements \cite{Shi2005Encyclopedia}.

The Ware label identifies the specific kiln sites where porcelains were produced. The dataset includes ten ware types from both northern and southern China, including Ding ware (Hebei), Ru and Jun ware (Henan), Peng ware (Shanxi), and Longquan ware (Zhejiang), along with Linchuan, Xianghu, and Jizhou wares from Jiangxi \cite{China1980}. It also includes Ge ware, whose precise production site remains unknown. These selections were based on their historical significance and dataset availability, particularly the Five Great Wares—Ding, Ru, Jun, Guan, and Ge—renowned for their influence on Chinese porcelain \cite{rastelli2016concept}. Each ware has unique characteristics: Ding ware is known for its thin, translucent white glaze \cite{rawson1984chinese}, Jun ware features blue-to-purple hues with occasional white suffusions \cite{medley1976chinese}, and Ge ware exhibits crackled glazes ranging from cream to greyish tones \cite{vainker1991chinese}. Ru ware is distinct for its pale 'duck-egg' blue glaze, sometimes extending into celadon green hues \cite{ru2017}.

The Glaze label classifies porcelains based on visual and textural properties such as color, brightness, transparency, and air bubble distribution. Experts traditionally identify glazes by evaluating these characteristics, as mineral pigments influence the final hue. Even transparent glazes exhibit subtle color variations, including moon-white, bluish-white, and yellowish-green. Notably, completely colorless transparent glazes are absent in archaeological porcelains \cite{MaoXiaohu2016AncientCeramicsIdentification}. The dataset includes seven glaze colors, reflecting the continuation of Tang Sancai traditions in the Song dynasty. Song Sancai porcelains were primarily used for practical artifacts such as incense burners, pillows, plates, and washes, typically featuring yellow, green, white, and brown glazes \cite{MaoXiaohu2016AncientCeramicsIdentification}.

The Type label categorizes artifacts based on their function and form. Song-era porcelain production flourished to meet the diverse needs of society, leading to a wide variety of forms such as bowls, plates, vases, saucers, washers, stoves, and pillows \cite{1982HistoryOfChineseCeramics}. These objects were designed for specific uses, including eating (plates, bowls), drinking (cups, jars), storage (pots, containers), and decoration (vases) \cite{2015HistoryOfChineseCeramicsIdentification}. Craftsmanship in this period balanced practicality with aesthetics, using variations in contour and proportion to create distinct forms. For example, vases in the dataset exhibit various styles, such as jade pot spring vases, plum vases, flat-bellied vases, melon prism vases, and gourd vases \cite{1982HistoryOfChineseCeramics}. Most designs emphasize slender, elegant shapes, while others prioritize symmetry and stability. The application of edge-detection techniques in preprocessing helps highlight object contours, making structural analysis—such as identifying patterns, cracks, and shape variations—more effective. Notably, Ding white vases feature elongated, slender forms emphasizing minimalism, while Guan and Longquan celadon vases adopt more rounded designs focused on volume and symmetry.

\begin{figure}[h]
  \centering
  \includegraphics[width=\linewidth]{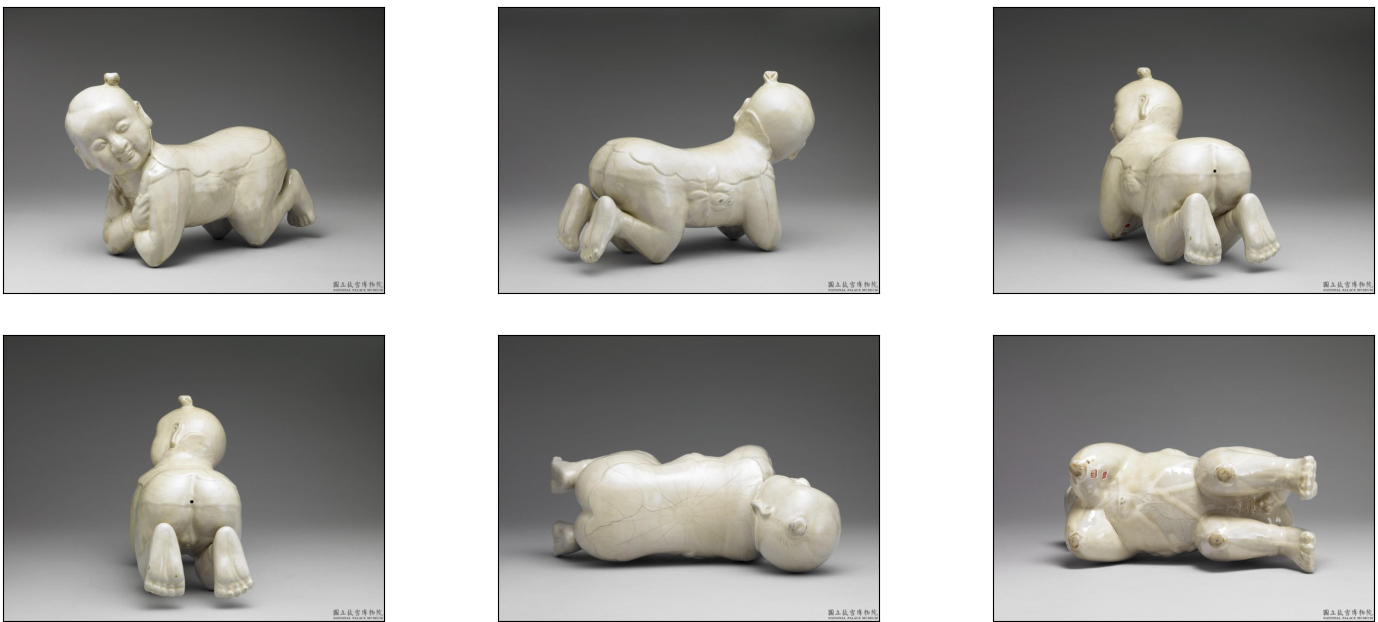}
  \caption{ Images of the same white porcelain pillow (Yuan dynasty, Peng ware) from multiple angles.}
  \label{fig:angle}
\end{figure}

Each porcelain of the dataset was photographed from different angles, as shown in Figure \ref{fig:angle}): front, left side, right side, back, top, and bottom views. These perspectives provide a comprehensive visual understanding of the porcelain objects.

\subsection{Multi-Task Learning Framework}

\begin{figure}[h]
\centering
\includegraphics[width=\linewidth]{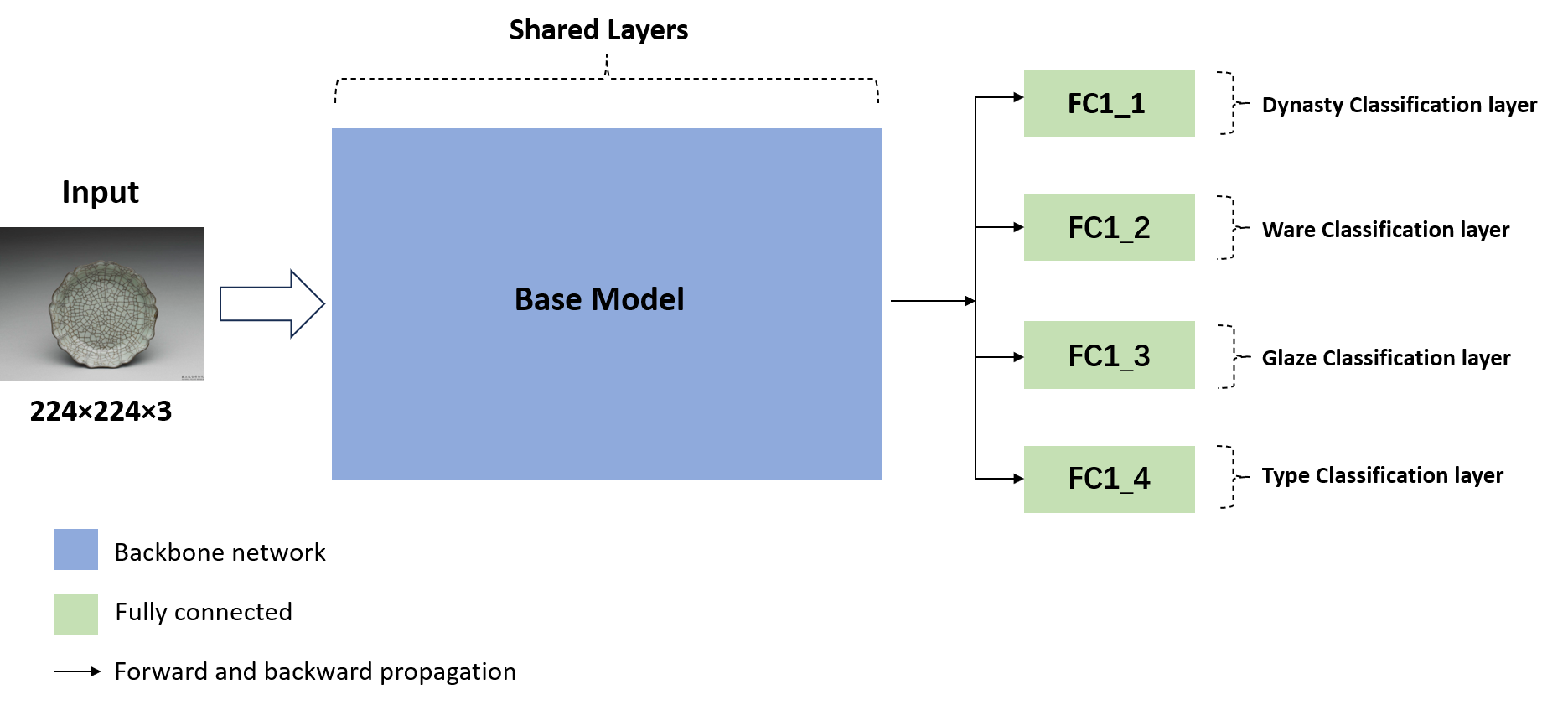}
\caption{Network Structure of the Multi-Task Model}
\label{fig:structure}
\end{figure}

The overall architecture to implement the multi-task learning framework is depicted in Figure~\ref{fig:structure}. The final classification layer of each base model is removed and replaced with four parallel fully connected layers, each corresponding to one classification task: dynasty, ware, glaze, and type. This structure allows the shared backbone to extract general features, while task-specific layers focus on individual classifications.  Adjustments were made based on the characteristics of each base model.

Multi-task learning offers several advantages by jointly learning related tasks \cite{zhang2018overview}:
\begin{itemize}
\item \textbf{Task Interdependence:} Porcelain features are often interrelated. Certain dynasties are associated with specific styles, glazes, and types of porcelain, meaning that features learned for ware, glaze, and type classification can aid in dynasty recognition. Similarly, different wares may have distinctive glazes and forms, while specific glaze types may be characteristic of particular dynasties or wares. This overlapping feature space enhances model performance across tasks.
\item \textbf{Shared Feature Learning:} The shared convolutional layers extract common low- and mid-level features such as edges, textures, and patterns, which are useful across all tasks. High-level features capture complex concepts like shapes, styles, and intricate patterns relevant to porcelain classification.
\end{itemize}

We took advantage of the performance of four widely used state-of-the-art CNN architectures:  
\begin{itemize}
\item \textbf{ResNet}: ResNet introduces “residual connections” that help mitigate the vanishing gradient problem, allowing deeper networks to train effectively \cite{he2016deep}. It has been extensively used in various computer vision applications. Among its different configurations (18, 34, 50, and 152 layers), we utilize the 50-layer variant (ResNet50).
\item \textbf{MobileNetV2}: Designed for efficiency, MobileNetV2 employs an "inverted residual structure" with residual connections between thin bottleneck layers, optimizing performance for mobile and embedded applications \cite{sandler2018mobilenetv2}.
\item \textbf{InceptionV3}: Developed by Google, InceptionV3 enhances computational efficiency by reducing the number of parameters while maintaining high accuracy. It uses multi-scale filters (1x1, 3x3, 5x5) in parallel within the same layer, enabling effective feature extraction across different scales \cite{szegedy2015going}. 
\item \textbf{Visual Geometry Group}: VGG deepens the network using small 3x3 convolutional filters while maintaining a simple and uniform design \cite{simonyan2014very}. Despite its relatively large number of parameters, it remains influential in computer vision. We adopt the 16-layer variant (VGG16) for this study.
\end{itemize}

All models employ transfer learning, leveraging pre-trained weights from the ImageNet dataset. Since images were all collected from the same museum, the shooting specifications, acquisition environment, and shooting conditions were basically the same for each piece of porcelain. Therefore, the only preprocessing methods used consisted of scaling them to a pixel size of 224x224 and data augmentation techniques (e.g., random horizontal flipping and rotation) applied after the train-test split.

\subsection{Loss Function and Evaluation Metrics}

The loss for each task is computed using the cross-entropy loss function, which is commonly used in classification problems, and measures the performance of a classification model whose output is a probability value between 0 and 1. It quantifies the difference between two probability distributions: the true distribution (ground truth labels) and the predicted distribution \cite{andreieva2020generalization}.
For a single example with $K$ classes, the cross-entropy loss $L$ is given by equation (\ref{equation1}):

\begin{equation}
L = -\sum_{i=1}^{K} y_i \log(p_i)
\label{equation1}
\end{equation}

$y_i$ - The ground truth probability for class $i$

$p_i$ - The predicted probability for class $i$

This loss function is appropriate for multi-class classification. In the training loop, the total loss is computed as the sum of the cross-entropy losses for each of the individual task outputs (dynasty, ware, glaze, type). The loss can be computed as equation (\ref{equation2}):

\begin{equation}
L_{\text{total}} = L_{\text{dynasty}} + L_{\text{ware}} + L_{\text{glaze}} + L_{\text{type}}
\label{equation2}
\end{equation}

Here, $L_{\text{dynasty}}$, $L_{\text{ware}}$, $L_{\text{glaze}}$ and $L_{\text{type}}$ represent the cross-entropy losses for the dynasty, ware, glaze, and type classification tasks, respectively.

Dynasty classification loss can be expressed as equation (\ref{equation3}):
\begin{equation}
L_{\text{dynasty}} = \text{CrossEntropyLoss}(\text{output}_{\text{dynasty}}, \text{label}_{\text{dynasty}})
\label{equation3}
\end{equation}

Ware classification loss can be expressed as equation (\ref{equation4}):
\begin{equation}
L_{\text{ware}} = \text{CrossEntropyLoss}(\text{output}_{\text{ware}}, \text{label}_{\text{ware}})
\label{equation4}
\end{equation}

Glaze classification loss can be expressed as equation (\ref{equation5}):
\begin{equation}
L_{\text{glaze}} = \text{CrossEntropyLoss}(\text{output}_{\text{glaze}}, \text{label}_{\text{glaze}})
\label{equation5}
\end{equation}

Type classification loss can be expressed as equation (\ref{equation6}):
\begin{equation}
L_{\text{type}} = \text{CrossEntropyLoss}(\text{output}_{\text{type}}, \text{label}_{\text{type}})
\label{equation6}
\end{equation}

To evaluate the performance in the various experiments, we employed the following metrics:  accuracy, precision, recall, and f1-score. The accuracy measures the ratio of correct predictions over the total number of samples evaluated. The precision is used to measure the positive label values that are correctly predicted from the total samples in a positive class, while the recall measures the fraction of positive label values that are correctly classified. Finally, the f1-score is computed as the harmonic mean of accuracy and recall. The corresponding formulas are reported in Equations \ref{equationAC}, \ref{equationPR}, \ref{equationRE}, and \ref{equationF1}.

\begin{equation}
\text{Accuracy} = \frac{TP + TN}{TP + TN + FP + FN}
\label{equationAC}
\end{equation}

Where: 

$TP$ - The positive sample with a positive predictive value 

$TN$ - The negative sample with a negative predictive value

$FP$ - The negative sample with a positive predictive value

$FN$ - The positive sample with a negative predictive value

\begin{equation}
\text{Precision} = \frac{TP}{TP + FP}
\label{equationPR}
\end{equation}

\begin{equation}
\text{Recall} = \frac{TP}{TP + FN}
\label{equationRE}
\end{equation}

\begin{equation}
\text{F1 Score} = 2 \times \frac{\text{Precision} \times \text{Recall}}{\text{Precision} + \text{Recall}}
\label{equationF1}
\end{equation}

\subsection{Experimental Settings}

This test platform utilized a server equipped with the following hardware and software specifications:
\begin{itemize}
    \item CPU: An AMD Ryzen 7 3700X, an 8-core processor. 
    \item RAM: 16 GB. 
    \item GPU: An NVIDIA GeForce RTX 2070 SUPER with 8 GB of dedicated video memory (VRAM).
    \item OS: Microsoft Windows 11. 
    \item Programming Language: Python 3.11. 
    \item DL Framework: PyTorch 2.1.2+cpu. This framework was employed for implementing and training DL models. 
    \item Computer Vision Library: Pillow 10.0.1. This library was used for image processing and manipulation tasks.
\end{itemize}

In terms of experimental parameters, the number of epochs is set to 50, and the batch size is set to 32. The Adam optimization algorithm is used to optimize the loss function. The learning rate is set to 0.001. The best version of the model, as determined by the lowest loss, is saved during training. At the same time, this study used the transfer learning technology, taking four models trained on the ImageNet dataset as the pre-trained network, and transferring its parameters as the starting point of training. To satisfy the conditions of the experiment, we randomly split the 5993 images into three non-overlapping subsets in an 8:1:1 ratio, with 80\% serving as the train set, 10\% as the valid set and 10\% as the test set.

\section{Results and Discussion}
\label{sec:res}

This Section presents the outcomes of our experiments and provides a comprehensive discussion of their implications. We begin with an analysis of the learning curves, offering insights into the training process and model convergence. Subsequently, we address each of our two research questions in dedicated subsections, presenting the relevant results and their interpretation.

\subsection{Training Analysis}

\begin{figure}[h]
\centering
\includegraphics[width=\linewidth]{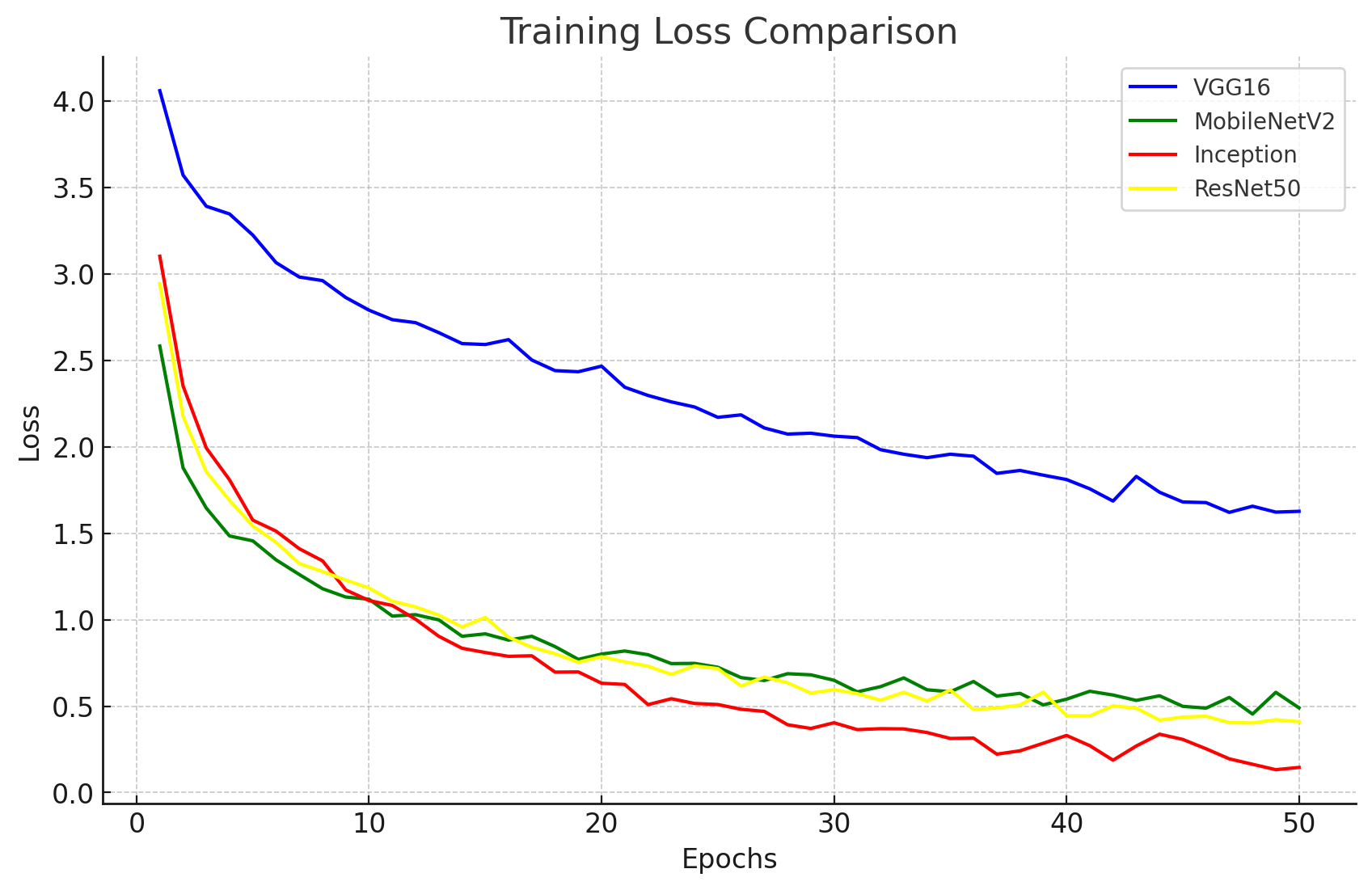}
\caption{Training Loss Comparison}
\label{fig:trainloss}
\end{figure}

To evaluate the performance of ResNet50, MobileNetV2, InceptionV3, and VGG16 on the porcelain dataset, we first analyzed their learning curves during joint training across the four classification tasks. The training loss progression for each model is shown in Figure~\ref{fig:trainloss}.

Among the four models, VGG16 exhibits the highest initial loss, approximately 4.0, indicating significant error at the start of training. In contrast, InceptionV3 and ResNet50 begin with lower initial loss values, around 3.1 and 3.0, respectively. MobileNetV2 demonstrates the lowest initial loss, approximately 2.5, suggesting a more favourable initialization.

During the first 10 epochs, all models undergo a rapid decrease in loss, demonstrating efficient parameter updates and fast convergence in the early training stages. MobileNetV2, InceptionV3, and ResNet50 exhibit the steepest decline in loss during this phase, indicating faster initial convergence compared to VGG16.

At later stages of training, InceptionV3 achieve the lowest final loss values, indicating superior learning performance. Anyway, MobileNetV2 and ResNet50, reached a comparable final loss level as InceptionV3. VGG16, regardless of continuous improvement, maintains the highest final loss, suggesting that it may require more training epochs or further hyperparameter tuning to enhance its performance.

Overall, InceptionV3, MobileNetV2, and ResNet50 demonstrate the most efficient convergence and lower final training loss, making them promising candidates for porcelain classification tasks.

\subsection{Multi-Task Classification of Porcelain Artifacts}

\begin{figure}[h]
\centering
\includegraphics[width=\linewidth]{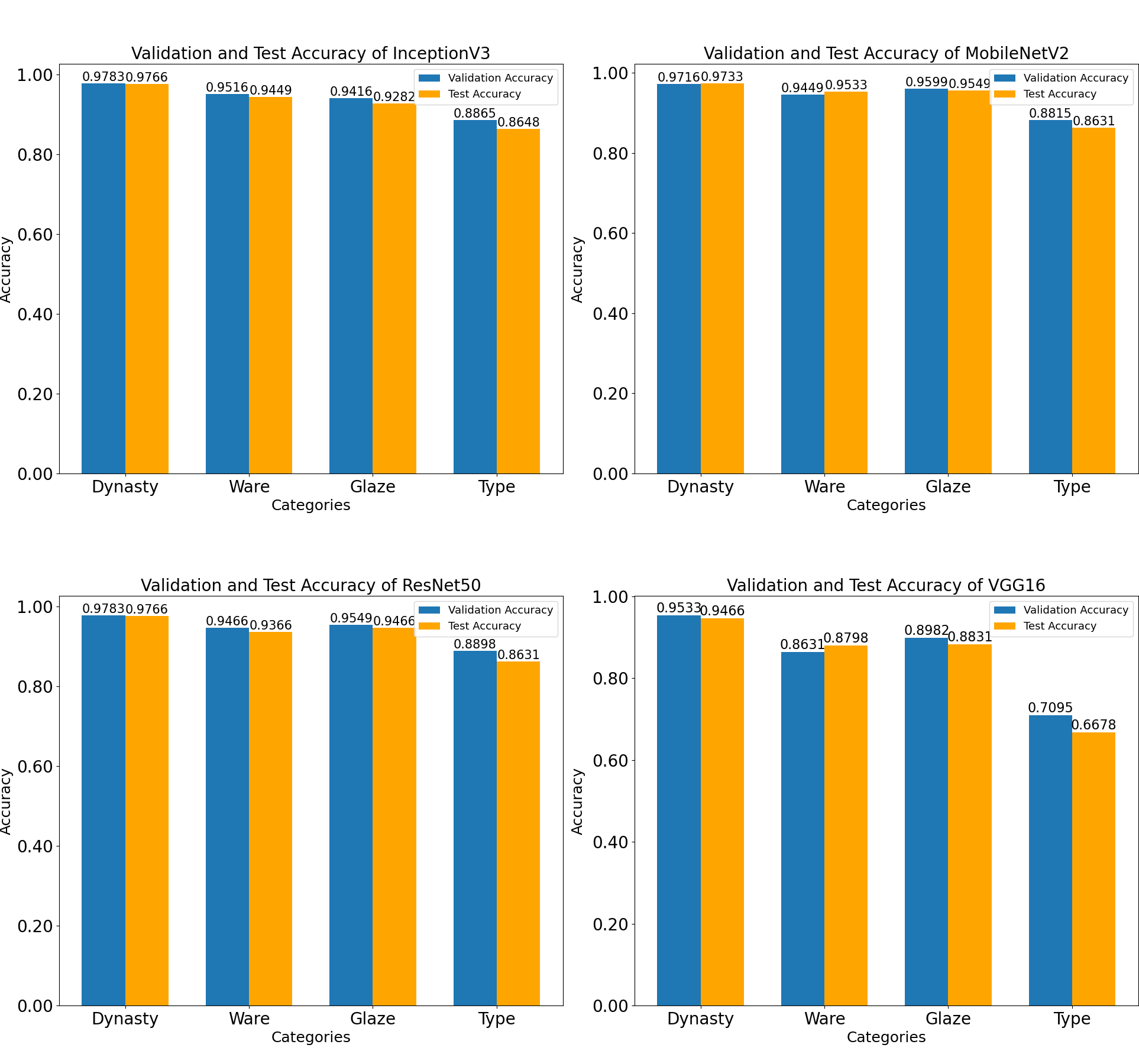}
\caption{Accuracy of Porcelain Classification Across Four Models}
\label{fig:accuracy}
\end{figure}

The classification accuracy of each model is evaluated based on the best-performing version saved during training. The results of the accuracy for the porcelain classification in the four models are presented in Figure~\ref{fig:accuracy}. 

The performance of each model is analyzed in four categories: Dynasty, Ware, Glaze, and Type. InceptionV3 maintains high accuracy across all categories, with its best performance in the Dynasty classification. ResNet50 also demonstrates strong accuracy, particularly in the Dynasty category, but experiences a slight drop in test accuracy compared to validation accuracy. MobileNetV2 exhibits consistent performance between validation and test sets, with minor improvements in the Ware classification. Meanwhile, VGG16 performs the worst overall, particularly struggling with the Type category, where it shows a significant accuracy drop. 


To further analyze classification performance, the confusion matrices are used to visualize the misclassification patterns  of each model across different tasks. 

\begin{figure}[h]
\centering
\includegraphics[width=\linewidth]{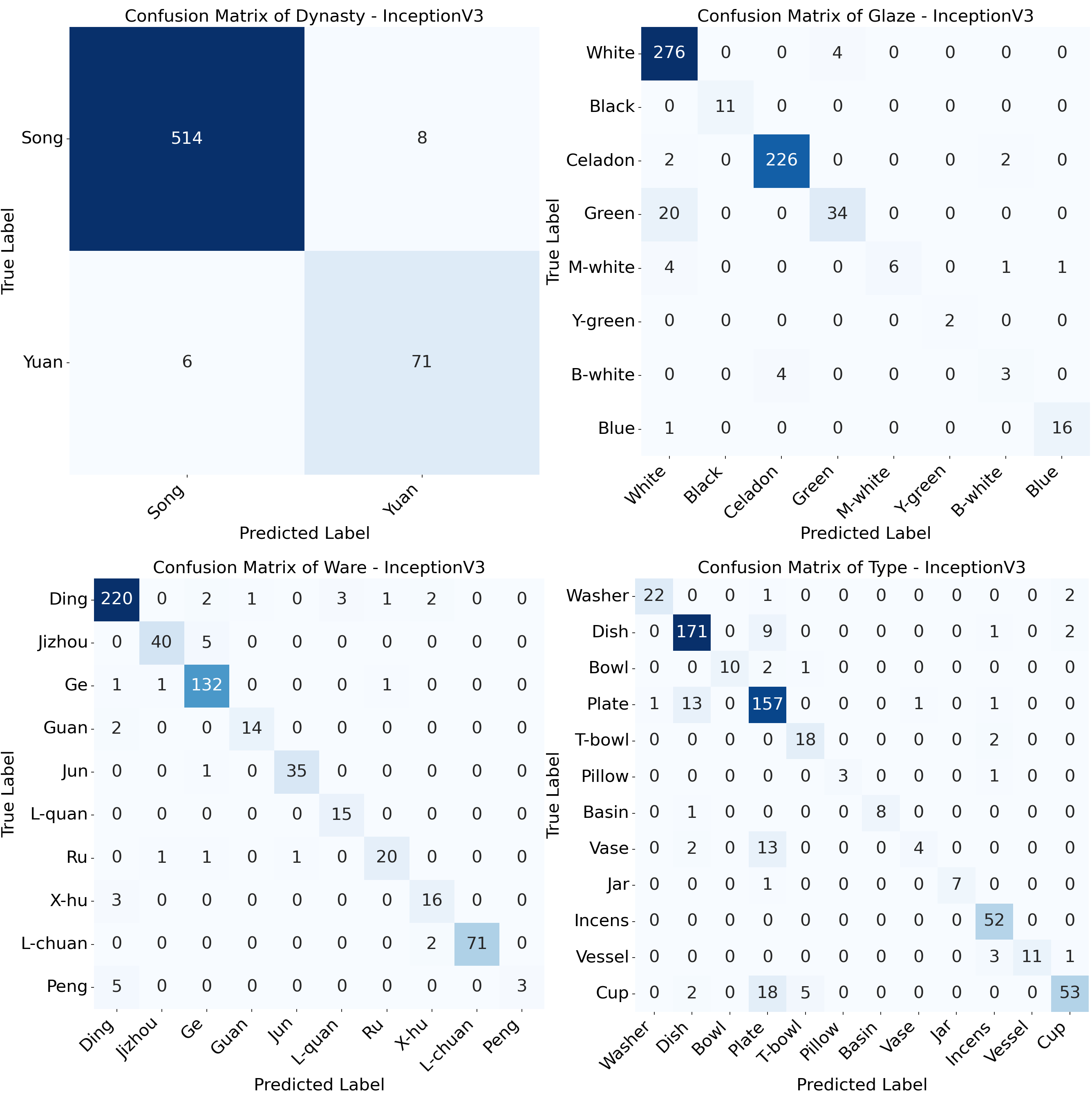}
\caption{Confusion Matrix of Four Tasks in InceptionV3}
\label{tab:cominception}
\end{figure}

The confusion matrices for each task using the InceptionV3 model are displayed in Figure~\ref{tab:cominception}. The model achieves high accuracy in classifying the Song and Yuan dynasties. For Glaze, it performs well in classes such as White and Celadon, which are the most represented ones, while it struggles with distinguishing Green. In the Ware category, high accuracy is observed in Ding and Ge classifications, but the model faces difficulties in correctly identifying Peng. In the Type category, the model shows strong classification for Dish and Plate but struggles with distinguishing Vase. 

\begin{figure}[h]
\centering
\includegraphics[width=\linewidth]{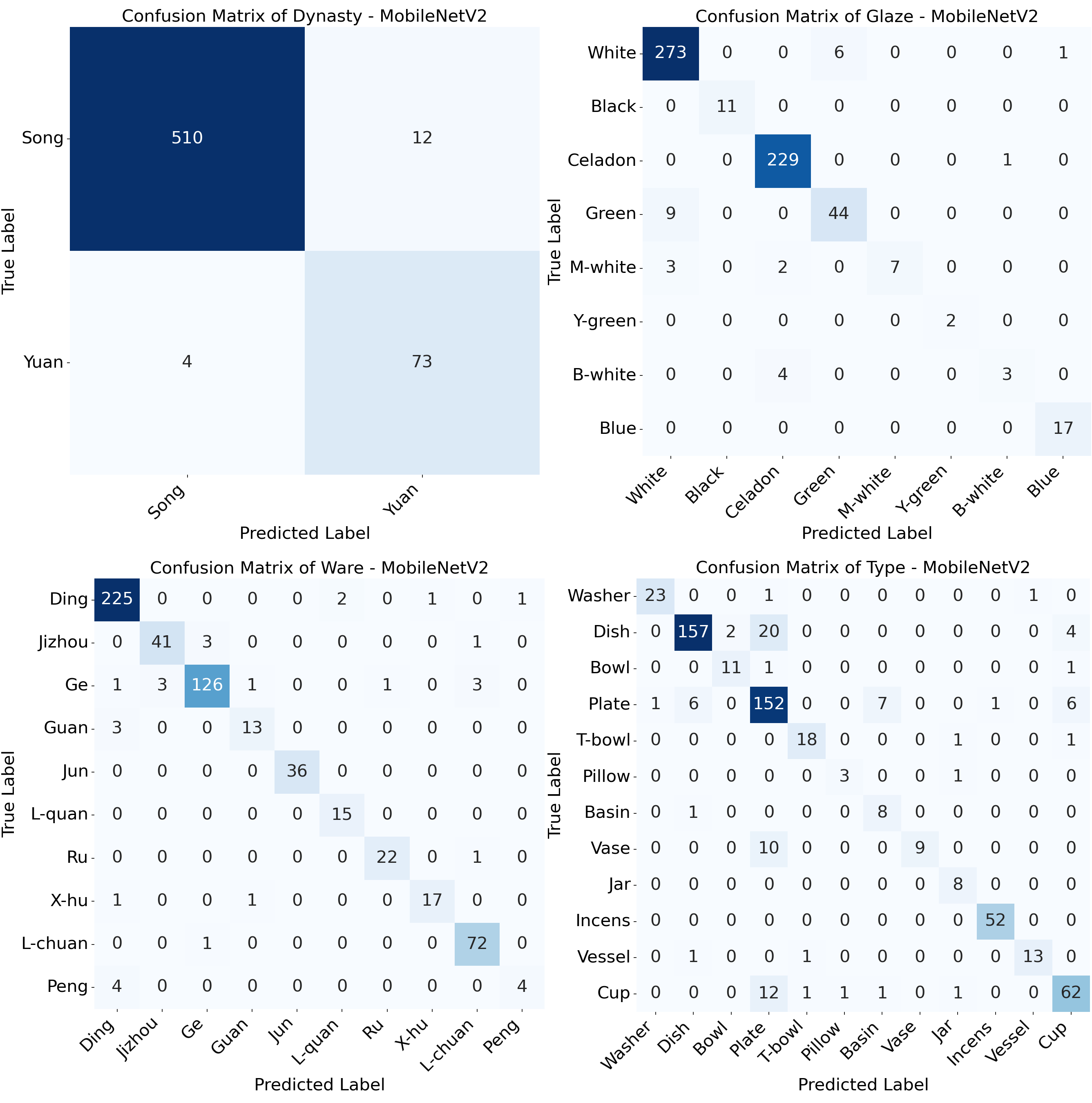}
\caption{Confusion Matrix of Four Tasks in MobileNetV2}
\label{tab:commobilnet}
\end{figure}

Similarly, the confusion matrices for MobileNetV2 are presented in Figure~\ref{tab:commobilnet}. The model accurately classifies the Song and Yuan dynasties and performs well in White and Celadon glaze categories. However, it struggles with distinguishing Peng in the Ware classification. MobileNetV2 shows slight improvements over InceptionV3 in certain tasks, demonstrating strong generalization.

\begin{figure}[h]
\centering
\includegraphics[width=\linewidth]{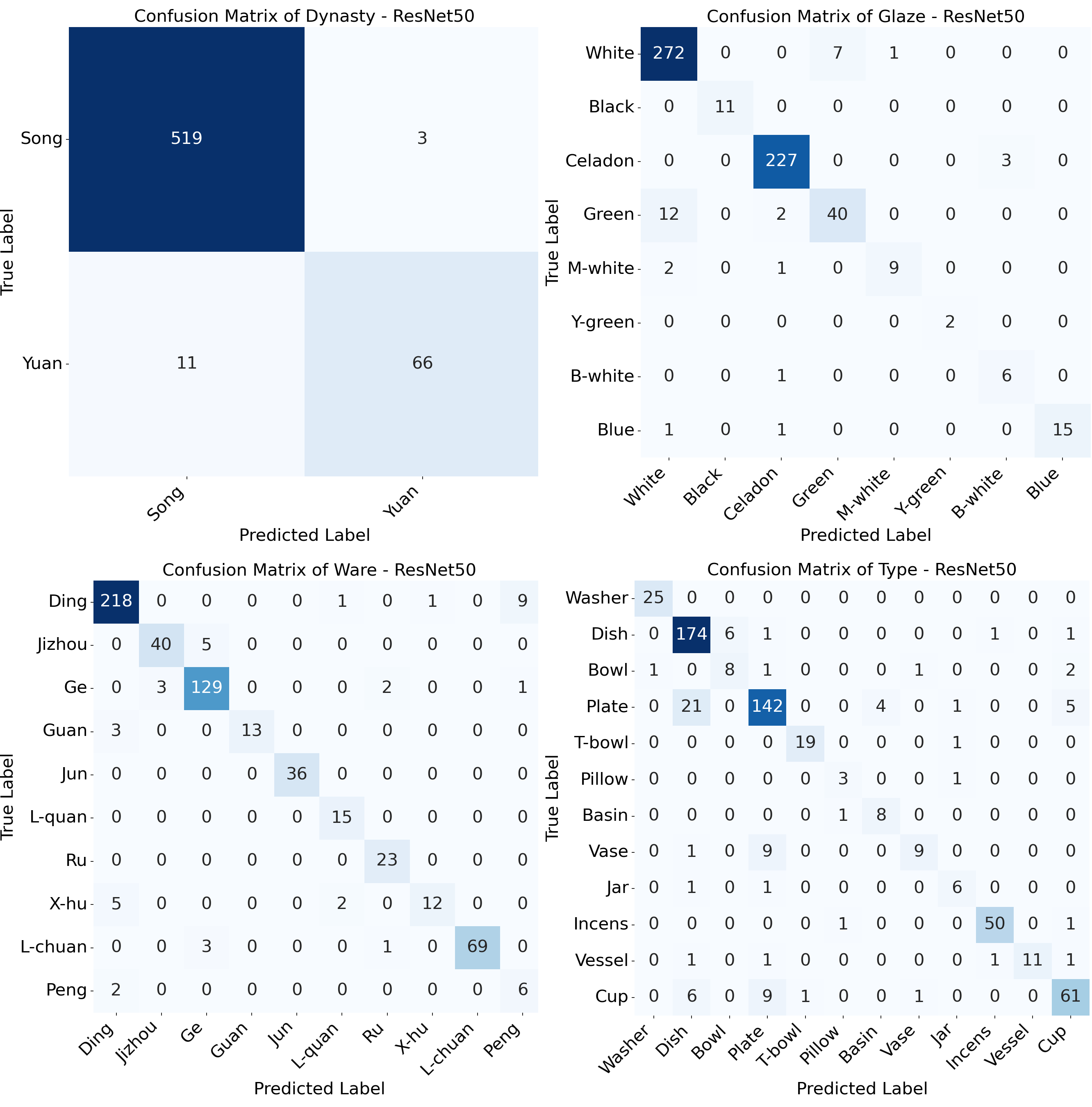}
\caption{Confusion Matrix of Four Tasks in ResNet50}
\label{tab:comresnet}
\end{figure}

For ResNet50, the confusion matrices are depicted in Figure~\ref{tab:comresnet} reveal high accuracy in the Dynasty classification, particularly for Song. However, the Yuan dynasty exhibits slightly more misclassifications. The model performs well in Glaze categories such as White and Celadon, while in Ware classification, it struggles to distinguish Peng. In Type classification, Dish and Plate are accurately classified, but Vase presents a challenge. Compared to MobileNetV2 and InceptionV3, ResNet50 demonstrates the most stable and accurate predictions.

\begin{figure}[h]
\centering
\includegraphics[width=\linewidth]{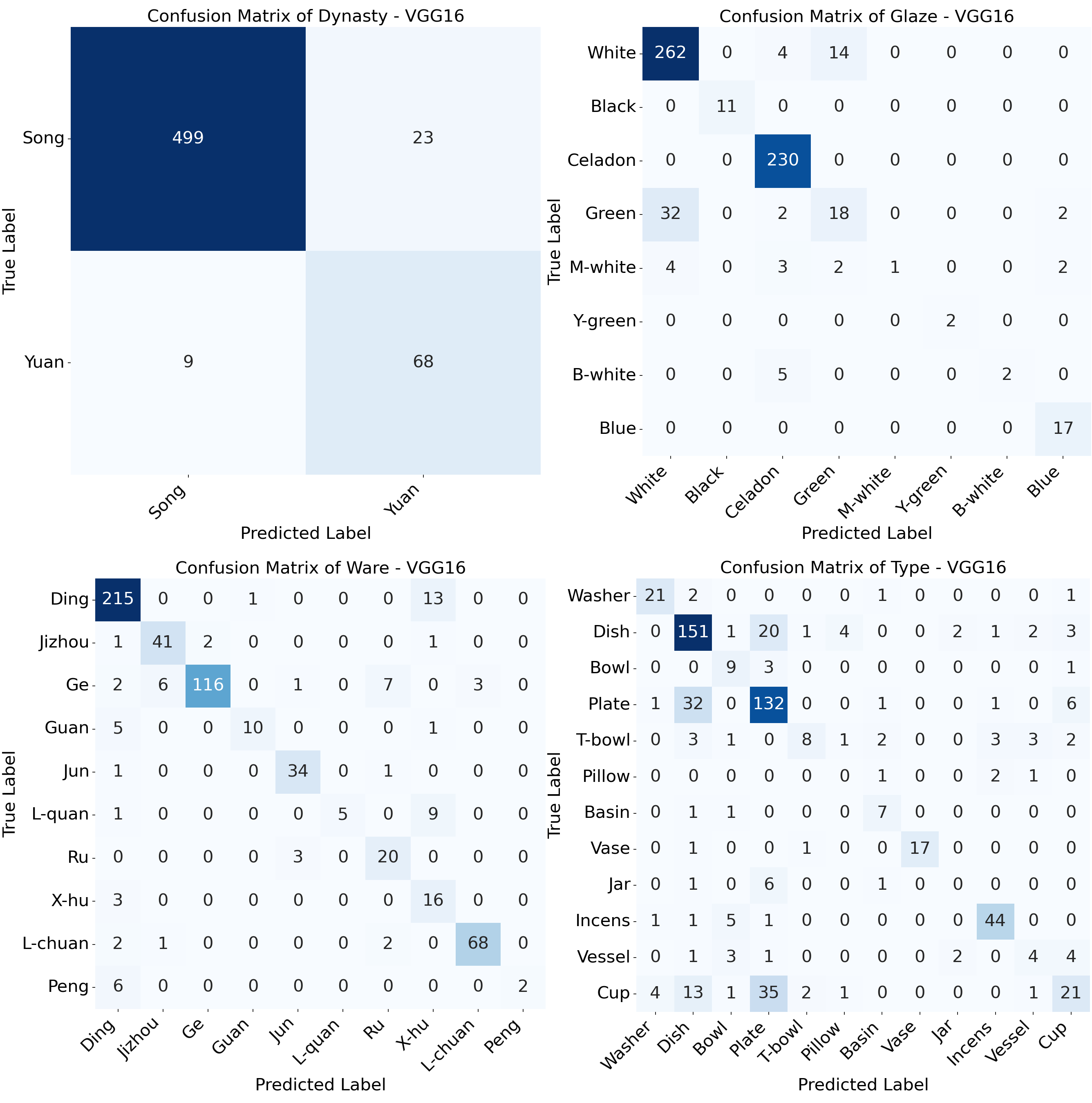}
\caption{Confusion Matrix of Four Tasks in VGG16}
\label{tab:comvgg}
\end{figure}

The confusion matrices for VGG16, shown in Figure~\ref{tab:comvgg}, indicate relatively high accuracy in the Song dynasty classification, though misclassification in the Yuan dynasty remains an issue. The model performs well for White and Celadon glazes but struggles with Green and Moonwhite. It also shows difficulties in distinguishing Longquan and Peng in the Ware category. Type classification presents the most challenges, with frequent misclassification between Cup, Pillow, Vase, and Jar. Overall, VGG16 has the weakest performance compared to the other models, exhibiting high misclassification rates.

\begin{figure}[h]
  \centering
  \includegraphics[width=\linewidth]{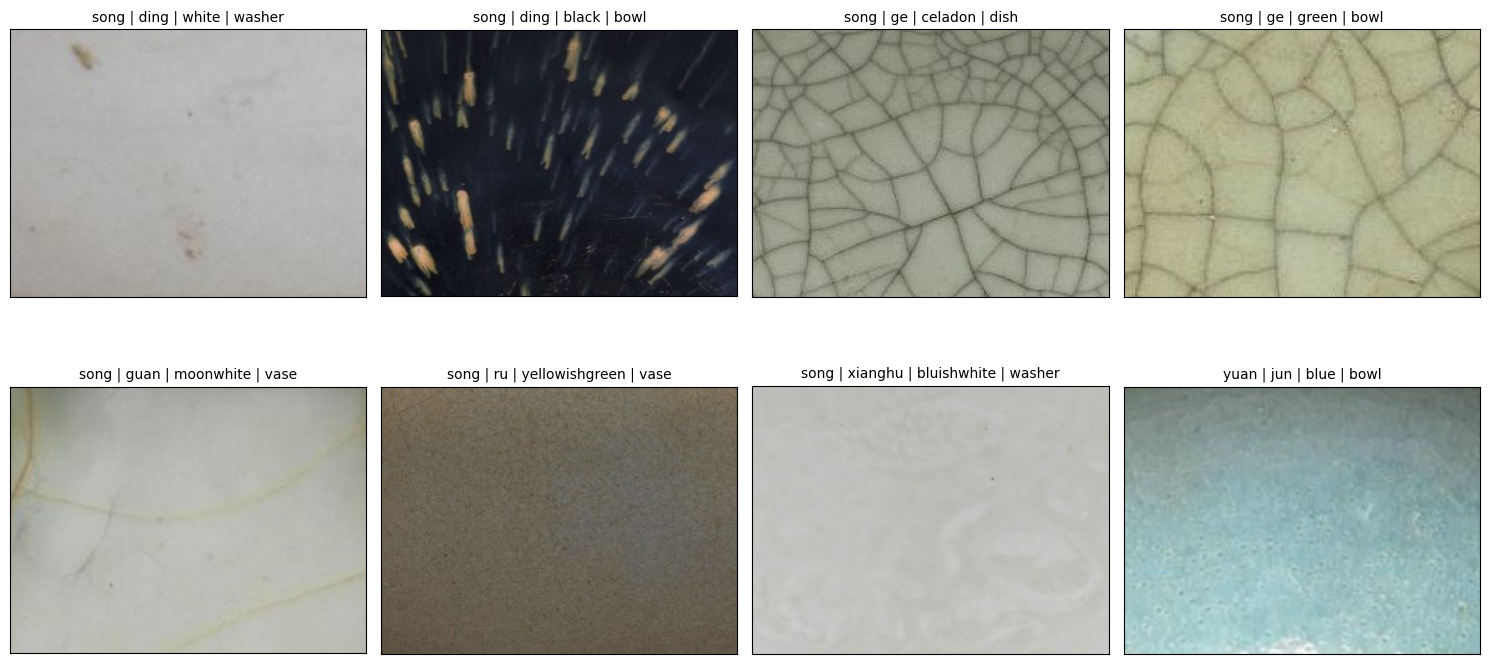}
  \caption{ Images of different glaze in the dataset }
  \label{fig:glaze}
\end{figure}

Further analysis of misclassification trends highlights specific challenges across all models. In the Ware task, MobileNetV2, InceptionV3, and VGG16 frequently misclassify Peng as Ding, likely due to their similar appearance as white porcelain. In the Glaze task, InceptionV3 and VGG16 struggle to classify Green correctly, often misidentifying it as White. In fact, Green lies between Celadon and White in hue, making classification difficult, as shown also in Figure \ref{fig:glaze}. The most significant challenge arises in the Type classification, where all models struggle to differentiate Vase from Plate. This misclassification is influenced by the dataset's inclusion of multiple viewing angles, especially top-down perspectives, which emphasize shape similarities between the two categories (Figure~\ref{fig:angle}). A common pattern across all models is that higher accuracy is consistently achieved for more represented classes, while less frequent or minority classes suffer from lower performance. This trend emphasizes the issue of data representation and availability, highlighting the importance of balanced datasets to ensure equitable classification performance across all categories.

To comprehensively compare model performance, Table~\ref{tab:comparison} presents validation accuracy, test accuracy, balanced accuracy, precision, recall, and F1-score obtained by each model focusing on the specific task. Given the dataset’s class imbalance (as shown in Table \ref{tab:labels_distribution}), balanced accuracy is used as a key metric to account for varying class sizes and provide a more standardized evaluation \cite{murphy2022probabilistic}.

\begin{table*}[h]
\centering
\caption{Comparison of Task-Specific Metrics for Multi-Task Models}
\label{tab:comparison}%
\begin{tabular}{|p{1.8cm}|p{1.0 cm}|p{1.5cm}|p{1.2cm}|p{1.2cm}|p{1.1cm}|p{1.0cm}|p{1.0cm}|}

\hline
 Model &Task & Validation set accuracy(\%)& Test set accuracy(\%)&Test set balanced accuracy(\%) &Precision & Recall & F1 Score\\
\hline
InceptionV3& Dynasty & 97.9 & 97.6 & 95.3 &0.976 & 0.976 & 0.976 \\
\hline
InceptionV3& Ware & 95.1 & 94.4 &87.3 &0.946 & 0.944 & 0.943 \\
\hline
InceptionV3& Glaze & 93.9 & 92.8 &77.0 &0.928 & 0.928 & 0.922 \\
\hline
InceptionV3& Type & 88.6 & 86.1 &79.4  &0.868 & 0.861 & 0.853 \\
\hline
MobileNetV2& Dynasty & 97.3 & 97.3 &96.2 &0.975 & 0.973 & 0.973\\
\hline
MobileNetV2&Ware & 94.4 & 95.3 & 89.7 &0.952 & 0.953 & 0.952\\
\hline
MobileNetV2& Glaze & 96.1 & 95.3 & 82.7 &0.952 & 0.953 & 0.952\\
\hline
MobileNetV2& Type & 87.9 & 86.1 &84.8 &0.866 & 0.861 & 0.861\\
\hline
ResNet50& Dynasty & 97.8 & 97.6 & 92.5 &0.976 & 0.976 & 0.976 \\
\hline
ResNet50& Ware & 94.6 & 93.0 & 89.3 &0.945 & 0.936 & 0.938 \\
\hline
ResNet50& Glaze & 95.4 & 94.8 & 88.3 &0.948 & 0.948 & 0.947 \\
\hline
ResNet50& Type & 89.1 & 86.1 & 80.6 &0.861 & 0.861 & 0.858 \\
\hline
VGG16& Dynasty & 95.3 & 94.6 &91.9 &0.952 & 0.946 & 0.948\\
\hline
VGG16& Ware & 86.1 & 87.9 & 75.0 &0.905 &0.879 & 0.880\\
\hline
VGG16& Glaze & 89.6 & 88.3 & 66.0 &0.874 & 0.883 & 0.865\\
\hline
VGG16& Type & 70.7 & 66.4 & 49.4 &0.639 & 0.664 & 0.635\\
\hline
\end{tabular}
\end{table*}

Across all tasks, MobileNetV2 consistently achieves high performance, maintaining strong accuracy, precision, recall, and F1 scores. It demonstrates robustness, especially in more challenging tasks like Ware and Type. InceptionV3 also performs remarkably well, particularly in Dynasty classification, though it shows slight performance degradation in more complex tasks like Type. ResNet50 follows closely, displaying balanced, high performance across most tasks. VGG16, however, exhibits a noticeable performance gap, especially in tasks requiring more nuanced feature differentiation, such as Glaze and Type.

All models excel in the Dynasty classification, with InceptionV3, MobileNetV2, and ResNet50 surpassing 97\% test accuracy and achieving near-perfect precision, recall, and F1 scores. MobileNetV2 slightly outperforms others in balanced accuracy (96.2\%), indicating strong generalization to less-represented classes. VGG16, while achieving respectable accuracy (94.6\%), lags behind the top models, suggesting that its feature extraction may be less refined for this task.

MobileNetV2 emerges as the top performer for Ware classification, achieving a 95.3\% test accuracy and balanced accuracy of 89.7\%, demonstrating resilience even with class imbalances. InceptionV3 and ResNet50 also perform well, with test accuracies of 94.4\% and 93.0\%, respectively. VGG16 struggles more notably here, with a test accuracy of 87.9\% and balanced accuracy dropping to 75\%, indicating potential misclassification issues, particularly for visually similar wares like Peng and Ding.

The Glaze task introduces greater challenges, with balanced accuracies dropping across all models. MobileNetV2 leads with a test accuracy of 95.3\% but still achieves only 82.7\% balanced accuracy, reflecting the difficulty in distinguishing minority glaze types such as Green. ResNet50 follows closely, achieving 94.8\% test accuracy and an improved balanced accuracy of 88.3\%, showing better handling of underrepresented glazes. InceptionV3, while achieving a strong test accuracy of 92.8\%, suffers from a lower balanced accuracy (77.0\%), indicating struggles with the same minority classes. VGG16, however, faces the greatest difficulty here, with its balanced accuracy dropping to 66.0\%, highlighting the model's weaker performance in nuanced glaze differentiation.

The Type classification task proves to be the most difficult for all models. MobileNetV2 and InceptionV3 maintain relatively strong performances, with balanced accuracies of 84.8\% and 79.4\%, respectively. ResNet50 achieves a balanced accuracy of 80.6\%, showcasing resilience despite the task’s complexity. VGG16, however, performs poorly, with its balanced accuracy plummeting to 49.4\% and an F1 score of 0.635, suggesting significant difficulty distinguishing between similar types, particularly the Vase and Plate categories. This disparity underscores VGG16’s limitations in handling diverse shape representations. The complexity of the Type classification task is further emphasized by the considerable diversity within porcelain types. The vase category alone encompasses more than ten distinct shapes from the Song dynasty, based purely on variations in form. Given that the dataset contains twelve different porcelain types, this diversity highlights the inherent challenges of accurately categorizing and analyzing porcelain artifacts in this task.

In response to the first research question \textbf{RQ1}, the results indicate that MobileNetV2 and ResNet50 emerge as the most robust models, demonstrating high accuracy and strong generalization across validation and test sets. InceptionV3 also performs well but shows minor overfitting. VGG16, instead, struggles significantly, particularly in distinguishing porcelain types. These findings suggest that MobileNetV2 is the most effective model for multi-task classification of porcelain artifacts, offering the best balance of accuracy and generalization.

\subsection{Impact of Transfer Learning on Porcelain Classification Accuracy}

To assess the impact of transfer learning, we conducted a comparative analysis of the four CNN models (MobileNetV2, InceptionV3, ResNet50, and VGG16) trained with and without pre-trained weights. Performance was evaluated based on training loss trends, validation accuracy, and multiple test set metrics, including accuracy, balanced accuracy, precision, recall, and F1 score.

\begin{figure}[h]
  \centering
  \includegraphics[width=\linewidth]{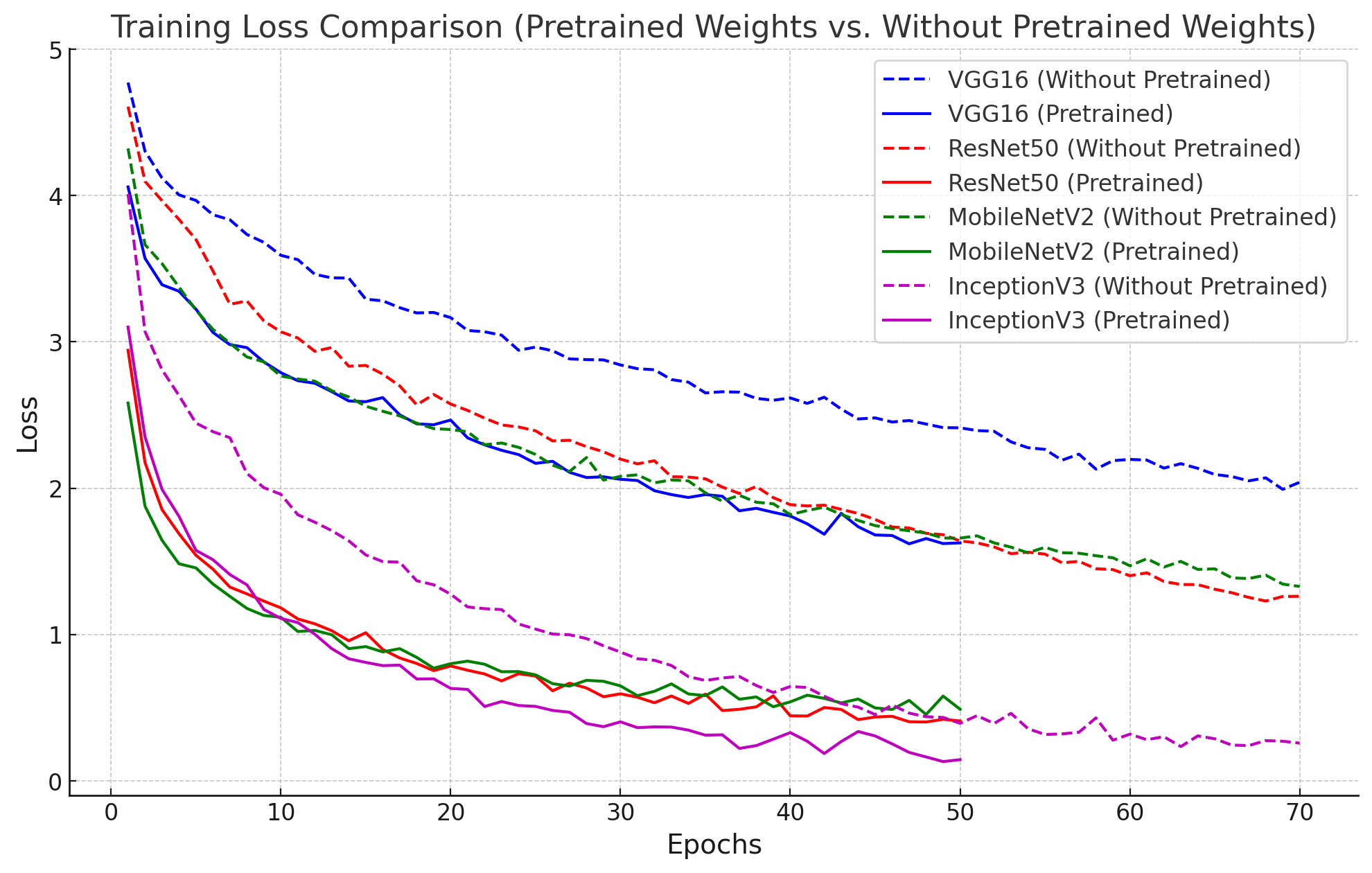}
  \caption{ Comparison of the training loss among pretrained and not pre-trained ones. }
  \label{fig:loss_transfer}
\end{figure}

Figure \ref{fig:loss_transfer} illustrates the training loss curves for models trained with and without transfer learning. A clear distinction can be observed: models initialized with pre-trained weights exhibit significantly faster convergence and lower overall training loss compared to those trained from scratch. This behavior aligns with the well-documented advantages of transfer learning, where pre-trained models require fewer epochs to reach optimal performance, reducing the risk of overfitting \cite{yosinski2014transferable}.

Table \ref{tab:comparison_transfer} presents a detailed comparison of performance metrics across different models and classification tasks, illustrating the impact of transfer learning. In all models and tasks, transfer learning consistently improves classification accuracy, balanced accuracy, and overall robustness. For example, MobileNetV2 demonstrates notable performance gains, with test accuracy improving from 73.3\% to 86.1\% for the "Type" task and balanced accuracy rising from 66.9\% to 84.8\%. InceptionV3 also shows substantial improvements, particularly in the "Ware" task, where test accuracy increases from 86.3\% to 94.4\%, and balanced accuracy jumps from 83.0\% to 87.3\%. ResNet50 exhibits strong performance boosts, especially in the "Glaze" task, with balanced accuracy improving from 68.6\% to 88.3\%. Even VGG16, which generally underperforms compared to the other models, benefits from transfer learning, achieving higher test accuracy in tasks such as "Ware" (82.6\% to 87.9\%) and "Glaze" (86.8\% to 88.3\%). 

\begin{table*}[ht]
\centering
\caption{Evaluation of the impact of transfer learning}
\label{tab:comparison_transfer}
\begin{tabular}{|p{1.8cm}|p{1.0 cm}|p{1.2cm}|p{1.8cm}|p{2.1cm}|p{1.5cm}|p{1.1cm}|p{0.8cm}|p{1.1cm}|}
\hline
\textbf{Model} & \textbf{Task} & \textbf{Transfer Learning} & \textbf{Test Accuracy (\%)} & \textbf{Balanced Test   Accuracy (\%)} & \textbf{Precision} & \textbf{Recall} & \textbf{F1 Score} \\ \hline
InceptionV3    & Dynasty       & Yes                        & 97.9                        & 95.3                                   & 0.976              & 0.976           & 0.976             \\ \hline
InceptionV3    & Dynasty       & No                         & 92.8                        & 92.6                                   & 0.945              & 0.928           & 0.933             \\ \hline
InceptionV3    & Ware          & Yes                        & 94.4                        & 87.3                                   & 0.946              & 0.944           & 0.943             \\ \hline
InceptionV3    & Ware          & No                         & 86.3                        & 83.0                                   & 0.881              & 0.863           & 0.868             \\ \hline
InceptionV3    & Glaze         & Yes                        & 92.8                        & 77.0                                   & 0.928              & 0.928           & 0.922             \\ \hline
InceptionV3    & Glaze         & No                         & 91.0                        & 78.6                                   & 0.910              & 0.910           & 0.904             \\ \hline
InceptionV3    & Type          & Yes                        & 86.1                        & 79.4                                   & 0.868              & 0.861           & 0.853             \\ \hline
InceptionV3    & Type          & No                         & 82.0                        & 75.9                                   & 0.820              & 0.820           & 0.817             \\ \hline
MobileNetV2    & Dynasty       & Yes                        & 97.3                        & 96.2                                   & 0.975              & 0.973           & 0.973             \\ \hline
MobileNetV2    & Dynasty       & No                         & 96.5                        & 93.6                                   & 0.966              & 0.965           & 0.965             \\ \hline
MobileNetV2    & Ware          & Yes                        & 95.3                        & 89.7                                   & 0.952              & 0.953           & 0.952             \\ \hline
MobileNetV2    & Ware          & No                         & 87.0                        & 77.0                                   & 0.888              & 0.870           & 0.872             \\ \hline
MobileNetV2    & Glaze         & Yes                        & 95.3                        & 82.7                                   & 0.952              & 0.953           & 0.952             \\ \hline
MobileNetV2    & Glaze         & No                         & 87.0                        & 69.5                                   & 0.895              & 0.902           & 0.886             \\ \hline
MobileNetV2    & Type          & Yes                        & 86.1                        & 84.8                                   & 0.866              & 0.861           & 0.861             \\ \hline
MobileNetV2    & Type          & No                         & 73.3                        & 66.9                                   & 0.717              & 0.733           & 0.711             \\ \hline
ResNet50       & Dynasty       & Yes                        & 97.6                        & 92.5                                   & 0.976              & 0.976           & 0.976             \\ \hline
ResNet50       & Dynasty       & No                         & 95.2                        & 91.1                                   & 0.954              & 0.952           & 0.952             \\ \hline
ResNet50       & Ware          & Yes                        & 93.0                        & 89.3                                   & 0.945              & 0.936           & 0.938             \\ \hline
ResNet50       & Ware          & No                         & 85.3                        & 74.9                                   & 0.866              & 0.853           & 0.853             \\ \hline
ResNet50       & Glaze         & Yes                        & 94.8                        & 88.3                                   & 0.948              & 0.948           & 0.947             \\ \hline
ResNet50       & Glaze         & No                         & 90.2                        & 68.6                                   & 0.897              & 0.902           & 0.889             \\ \hline
ResNet50       & Type          & Yes                        & 86.1                        & 80.6                                   & 0.861              & 0.861           & 0.858             \\ \hline
ResNet50       & Type          & No                         & 75.0                        & 67.7                                   & 0.732              & 0.750           & 0.735             \\ \hline
VGG16          & Dynasty       & Yes                        & 94.6                        & 91.9                                   & 0.952              & 0.946           & 0.948             \\ \hline
VGG16          & Dynasty       & No                         & 95.2                        & 86.2                                   & 0.950              & 0.952           & 0.950             \\ \hline
VGG16          & Ware          & Yes                        & 87.9                        & 75.0                                   & 0.905              & 0.879           & 0.880             \\ \hline
VGG16          & Ware          & No                         & 82.6                        & 66.6                                   & 0.836              & 0.826           & 0.821             \\ \hline
VGG16          & Glaze         & Yes                        & 88.3                        & 66.0                                   & 0.874              & 0.883           & 0.865             \\ \hline
VGG16          & Glaze         & No                         & 86.8                        & 57.0                                   & 0.837              & 0.868           & 0.842             \\ \hline
VGG16          & Type          & Yes                        & 66.4                        & 49.4                                   & 0.639              & 0.664           & 0.635             \\ \hline
VGG16          & Type          & No                         & 64.9                        & 53.4                                   & 0.634              & 0.649           & 0.630             \\ \hline
\end{tabular}
\end{table*}

These findings align with existing research demonstrating the effectiveness of transfer learning in improving classification performance in specialized domains, including medical imaging \cite{raghu2019transfusion}, remote sensing \cite{tuia2016domain}, and cultural heritage preservation \cite{liu2023neural}. The improved performance can be attributed to the ability of pre-trained CNNs to capture generic visual features, such as edges, textures, and shapes, which are transferable to new tasks, especially when fine-tuned on domain-specific data.

In the context of porcelain artifact classification, transfer learning proves particularly valuable due to the intricate and diverse nature of ceramic features. As discussed in the previous Section, porcelain types exhibit considerable variation in shape, glaze, and decoration. By leveraging pre-trained feature extractors, CNNs can better distinguish such fine-grained differences, resulting in superior classification performance.

In conclusion, to answer \textbf{RQ2}, our results confirm that transfer learning significantly enhances the classification of porcelain artifacts across dynasty, glaze, ware, and type tasks. The use of pre-trained CNNs not only accelerates convergence and stabilizes training but also improves classification accuracy and robustness, making it a highly effective approach for tasks with limited domain-specific training data.

\subsection{Limitations}

While this study demonstrates the potential of multi-task learning models for porcelain classification, several limitations must be acknowledged.

Firstly, class imbalance remains a critical challenge. Despite efforts to balance the dataset, certain categories — particularly within the Glaze and Type tasks — remain underrepresented, leading to performance degradation for these classes. This imbalance is reflected in lower balanced accuracies, especially for VGG16, and highlights how data availability influences model performance. Future work could explore advanced techniques like data augmentation, synthetic data generation, or class-specific loss functions to mitigate this issue.

Secondly, visual similarity between certain categories introduces ambiguity that current models struggle to resolve. In the Ware task, for example, models often misclassify Peng as Ding due to their similar white porcelain appearance. Similarly, the Green glaze is frequently misidentified as White or Celadon, reflecting how subtle hue differences challenge even high-performing architectures. This underscores the need for more sophisticated feature extraction methods, potentially integrating domain-specific knowledge or multi-modal inputs such as spectral imaging.

Another limitation is viewing angle variability, particularly in the Type classification task. The inclusion of multiple perspectives, including top-down views, sometimes obscures distinctive shape features, causing confusion between categories like Vase and Plate. Future datasets could benefit from standardized imaging protocols or 3D data to provide more comprehensive representations of complex objects.

Moreover, well-known CNN architectures — InceptionV3, MobileNetV2, ResNet50, and VGG16 — were employed in this study. While these models serve as strong, state-of-the-art baselines, they were chosen primarily to provide a performance benchmark rather than to propose novel architectures. Future research could explore more specialized or hybrid architectures tailored to the unique characteristics of porcelain classification, potentially integrating attention mechanisms or transformer-based vision models for enhanced feature extraction.

Finally, generalizability to unseen datasets remains uncertain. While the models perform well on the curated dataset used in this study, their robustness against artifacts from different collections, regions, or historical contexts is untested. Addressing this requires further validation on diverse datasets, ensuring the models maintain accuracy beyond the current domain.

\section{Conclusion and Future Works}
\label{sec:conc}

This study explored the performance of various CNN architectures — MobileNetV2, InceptionV3, ResNet50, and VGG16 — in classifying porcelain artifacts across four key attributes: dynasty, glaze, ware, and type. Through comprehensive evaluation, we demonstrated that transfer learning significantly improves model performance, enabling faster convergence, better generalization, and higher accuracy, particularly for complex classification tasks like type identification. Among the tested models, MobileNetV2, InceptionV3, and ResNet50 consistently outperformed VGG16, highlighting the importance of architecture design and pre-trained feature extraction in handling intricate visual datasets.

The results emphasize that transfer learning is especially beneficial in domains where acquiring large, labeled datasets is impractical — a common scenario in cultural heritage and artifact classification. The pre-trained models, having learned a rich set of generic visual features from large-scale datasets like ImageNet, proved more adept at extracting meaningful representations from the porcelain images. This capability is crucial for distinguishing subtle variations in shape, glaze, and decoration, which are essential for accurate classification.

Despite the promising results, some limitations remain, paving the way for future research. One key challenge lies in the "Type" classification task, where even the best-performing models struggled to achieve balanced accuracy comparable to other tasks. This indicates that further advancements in feature extraction and model specialization are necessary to handle such high intra-class variability.

Future research could focus on combining the strengths of different architectures, integrating lightweight models like MobileNetV2 with more complex designs such as Inception or ResNet to balance efficiency and performance. Enhancing these models with advanced attention mechanisms, such as Squeeze-and-Excitation \cite{hu2018squeeze} or Vision Transformers \cite{raghu2021vision}, could further improve feature extraction by guiding the model to focus on the most relevant parts of the artifact images. While pre-training on large datasets like ImageNet offers a strong foundation, this data is designed for general object recognition and may not fully capture the nuanced visual characteristics of porcelain artifacts. Developing a domain-specific pre-training dataset that incorporates historical artwork, ceramics, and other cultural artifacts could yield more specialized feature representations, particularly benefiting complex tasks like type classification, where models currently struggle. Data scarcity remains a challenge in cultural heritage domains, so leveraging advanced data augmentation techniques and synthetic data generation — potentially supported by generative adversarial networks (GANs) or Large Language Models — could help expand datasets with realistic, diverse samples. This would allow models to better generalize to underrepresented porcelain types. Furthermore, ensuring model transparency is crucial for encouraging adoption among cultural heritage experts. Integrating explainable AI techniques, such as Grad-CAM \cite{selvaraju2017grad}, could provide visual explanations of model predictions, offering valuable insight into the decision-making process and fostering trust in AI-driven analysis. Beyond porcelain artifacts, extending this approach to other cultural artifacts like paintings, sculptures, and textiles would test the adaptability of these models across domains, highlighting their broader potential in cultural heritage preservation. Additionally, investigating few-shot learning strategies could enhance model performance on unseen artifact categories, enabling effective classification with only minimal labeled data — a crucial advantage in resource-constrained domains where extensive annotation is impractical.

\section*{Acknowledgment}

The support provided by the China Scholarship Council (CSC) during the Ph.D. research of Ziyao Ling at the University of Bologna is acknowledged.

\bibliography{sn-bibliography}

\end{document}